\documentclass[journal]{IEEEtran}
\usepackage{setspace}
\ifCLASSINFOpdf
\else
\fi
\usepackage[dvipsnames]{xcolor}
\usepackage{color}
\usepackage{subcaption}
\usepackage{float}
\usepackage{graphicx}
\usepackage[left=0.75in,right=0.75in,bottom=0.75in,top=0.75in]{geometry}
\usepackage{amssymb}
\usepackage{amsmath}

\usepackage{cuted}
\usepackage{lipsum, array, booktabs}
\usepackage{tikz}
\usetikzlibrary{shapes.geometric, arrows}
\usepackage{tabularx,ragged2e}
\usepackage{booktabs}
\newcolumntype{Y}{>{\centering\arraybackslash}X}
\usepackage{mathtools}

\newcommand{\yi}{y_i}
\newcommand{\yihat}{\hat{y}_i}

\newcommand{\YT}{\mathrm{y}}

\newcommand{\XT}{\mathbf{X}}
\newcommand{\XTprime}{\mathbf{X}^{\top}}

\newcommand{\betat}{\boldsymbol{\beta}}
\newcommand{\betatnot}{\beta_{0}}
\newcommand{\betatnothat}{\hat{\beta}_0}

\newcommand{\betaT}{\boldsymbol{\beta}}
\newcommand{\betaThat}{\hat{\boldsymbol{\beta}}}
\newcommand{\betaTprime}{\boldsymbol{\beta}^{\top}}
\newcommand{\betaTprimehat}{{\hat{\boldsymbol{\beta}}{}^{\top}}}

\newcommand{\betatkprimehat}{\hat{\boldsymbol{\beta}}{}^{p\delta}{}^{\top}}
\newcommand{\XmT}{\mathbf{x}_{i}}

\newcommand{\qdriving}{q_{i}^1}
\newcommand{\qtransit}{q_{i}^0}

\newcommand{\taum}[1]{\boldsymbol{\tau}_i^{#1}}

\begin{document}

\title{\vspace{0.25in}Interpretable Machine Learning Models for Modal Split Prediction in Transportation Systems}




\author{Aron Brenner, Manxi Wu, and Saurabh Amin
\thanks{
A. Brenner (abrenner@mit.edu) and S. Amin (amins@mit.edu) are with the Laboratory for Information and Decision Systems at the Massachusetts Institute of Technology; M. Wu (manxiwu@berkeley.edu) is with the Department of Electrical Engineering and Computer Sciences at the University of California at Berkeley}}

\markboth{Journal of \LaTeX\ Class Files,~Vol.~14, No.~8, August~2015}%
{Shell \MakeLowercase{\textit{et al.}}: Bare Demo of IEEEtran.cls for IEEE Journals}
%

\maketitle
\thispagestyle{empty}
\pagestyle{empty}

\begin{abstract}
Modal split prediction in transportation networks has the potential to support network operators in managing traffic congestion and improving transit service reliability. We focus on the problem of hourly prediction of the fraction of travelers choosing one mode of transportation over another using high-dimensional travel time data. We use logistic regression as base model and employ various regularization techniques for variable selection to prevent overfitting and resolve multicollinearity issues. Importantly, we interpret the prediction accuracy results with respect to the inherent variability of modal splits and travelers’ aggregate responsiveness to changes in travel time. By visualizing model parameters, we conclude that the subset of segments found important for predictive accuracy changes from hour-to-hour and include segments that are topologically central and/or highly congested. We apply our approach to the San Francisco Bay Area freeway and rapid transit network and demonstrate superior prediction accuracy and interpretability of our method compared to pre-specified variable selection methods.
\end{abstract}


\IEEEpeerreviewmaketitle

\section{Introduction}

\begin{figure}[t!]
    \centering
    \input{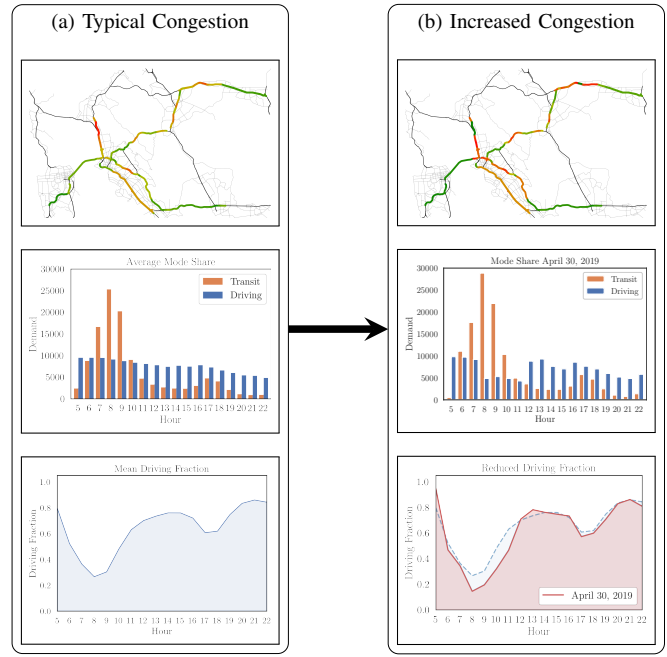}
    \caption{Increased congestion on the morning of April 30, 2019 resulted in a lower share of driving demand in the San Francisco Bay Area as travelers switched to public transit. Predictive information about the impact of such congestion events can potentially be useful in scheduling of rapid transit.}
    \label{fig:intro}
\end{figure}

Multi-modal transportation system operations depend crucially on the distribution of aggregate travel demand across travel modes and the way in which this modal split varies over time. Accurate modal split prediction at a fine temporal scale can support transportation authorities in better managing congestion and improving service reliability. This information can be especially valuable in highly uncertain situations when traveler preferences shift significantly from one mode to another. For example, accurate modal split prediction in times when travel demand changes from off-peak to peak periods can help transit operators in scheduling frequent service in high-demand corridors, which in turn can induce further demand shift to transit and reduce freeway congestion during peak periods (see Fig. \ref{fig:intro}). Additionally, improved understanding of how the modal split is shaped by spatio-temporal variations in travel times can provide guidance on interventions such as control \cite{jin2018} or tolling \cite{yan1996} of specific freeway segments. 

In this paper, we propose an interpretable machine learning (ML) approach to (1) predict the modal split in transportation systems on an hourly basis using transportation network travel times; and (2) identify the subset of network segments that are most critical for predictive accuracy. We adopt logistic regression as a base model and investigate the use of different variable selection (regularization) methods to tackle the high dimensionality of travel time data and identify segments that have the most impact in improving the predictive accuracy. We instantiate our approach on data from the San Francisco (SF) Bay Area, namely regions in the East Bay and SF proper, which are served by the Bay Area Rapid Transit (BART) and a congestion-prone freeway system.

Our approach directly predicts the hourly aggregate modal split based on raw, segment-level travel time data from a large number of regions where both transit and driving choices are available. Essentially, the logistic hypothesis captures mode choice behavior at an aggregate level while the associated variable selection and regularization methods control the variance of the predicted modal split. This makes our approach well-suited for generating modal split predictions on an hourly basis, in comparison to the classical approach in the travel demand literature that typically relies on survey data to build a discrete-choice model of traveler preferences \cite{benakiva1985} and estimates demand patterns through extensive micro-simulations \cite{benakiva1998, bhat2004}.

Importantly, the resulting models are interpretable in that the selected features correspond to segments impacting modal split predictions due to their location in the network as well as their influence on the overall congestion dynamics. Indeed, previous work has used specific criteria to pre-select segments in freeway networks based on their location or impact on congestion \cite{pu2012}, \cite{jayasinghe2015}. In contrast, we leverage recent developments in regularized regression \cite{friedman2010} and obtain superior predictive performance in comparison to pre-specified variable selection methods that only select segments based on their location in the network (betweenness centrality) or local congestion measures (segment-level speeds). Our computational results demonstrate the effectiveness of variable selection methods in tackling high-dimensional travel time data and identifying a relatively small subset of segments that are most critical for out-of-sample predictive accuracy.

Our paper contributes to the emerging literature that explore the use of ML models such as classification and regression trees (CART) and neural networks for discrete-choice modeling \cite{nijkamp, hensher, zegras2017, cheng2019, zhao2020}; for a more thorough discussion of applications of ML towards choice modeling, we refer readers to \cite{cranenburgh2022}. These approaches treat choice modeling as a classification problem and typically focus on out-of-sample predictive accuracy as the main performance measure. Thus, they naturally leverage the ability of ML models to arrive at the desired bias-variance tradeoff by minimizing overfitting. We go a step further by focusing on how variable selection methods can be used to not only build sparse models (similar to \cite{cheng2019, zegras2017}), but also to generate new insights by properly interpreting them.

The rest of the paper is organized as follows: In Sec. \ref{sec:problem}, we present our modal split prediction model and discuss the method of regularized maximum likelihood estimation as well as the model selection and evaluation process. In Sec. \ref{sec:interpretation}, we introduce pre-specified variable selection methods that serve as benchmarks for evaluating our approach and detail how the accuracy of our prediction model can be interpreted in terms of variability of modal split and responsiveness to changes in travel time. Finally in Sec. \ref{sec:empirical}, we apply our modeling and interpretation framework to a case study in the SF Bay Area.
\section{Modeling Approach}\label{sec:problem}
\subsection{Prediction Model}
In this section, we present our model for predicting in time interval $t$ the fraction of driving flow among the total flows of driving and transit based on travel time data for segments in a multi-modal transportation network. In our case study, each $t$ corresponds to a one-hour interval, although $t$ can vary in practice depending on the time granularity of available data. We estimate separate models for different times of day since travelers and their demand patterns can vary considerably throughout the day. For the observation corresponding to day $i$, we denote by $\qdriving$ and $\qtransit$ the driving flow and the transit ridership from region A to region B, respectively. Here, regions A and B correspond respectively to the sets of origin and destination locations that are served by both freeway and transit modes (Fig. \ref{fig:map}). We denote by $\yi \in [0,1]$ the driving fraction on day $i$:
\begin{align*}
    \yi=\frac{\qdriving}{\qdriving+\qtransit},
\end{align*}
which is the quantity we aim to predict.

Our multi-modal network is comprised of the freeway network and the transit network. We define segments in the freeway network ($n_{\text{driving}}$ segments in total) as sections between adjacent entrance or exit ramps and segments in the transit network ($n_{\text{transit}}$ segments in total) as sections between two adjacent stations. The total number of segments is $n=n_{\text{driving}} + n_{\text{transit}}$.

For a given time interval $t$ and day $i$, we predict the driving fraction $\yi$ using the travel time of all segments in the network during time intervals $t, (t-\delta), \dots, (t-p\delta)$, where $\delta>0$ denotes a unit time lag and $p$ the maximum number of time lags incorporated in our model. Our prediction model uses travel time vectors in lagged time intervals because the flows of driving and transit counted in any time interval are induced by travelers' mode choice decisions that may have been made before their trip was counted by the traffic sensor or transit tap-in sensor. As such, some mode choices made in one time interval will influence the flow count for a subsequent time interval. The unit time lag variable $\delta$ depends on the granularity of the available travel time data, and $p$ is chosen such that $p\delta$ is larger than the maximum travel time from any origin in region $A$ to the sensors that record the flow between $A$ and $B$ in the network. In our case study, we choose $p=6$ and $\delta=\text{10 minutes}$ -- i.e. one hour of travel time data in 10-minute intervals.

Thus, for each time interval $t$, our goal is to predict $\yihat$ using the travel time vector $(\taum{0}, \taum{\delta}, \dots, \taum{p\delta})^\top \in \mathbb{R}^{n(p+1)}$, where ${\taum{k\delta}=(\tau_{i,j}^{k\delta})_{j \in 1,\dots,n}}$ is the vector of travel times for all $n$ segments in time interval $(t-k\delta)$ on day $i$. We predict $\yihat$ as
\begin{align}\label{eq:logistic}
    \yihat &= \sigma(\betatnothat + \hat{\boldsymbol{\beta}}{}^0{}^{\top} \taum{0} + \dots + \betatkprimehat\taum{p\delta}) \notag \\
    &= \frac{1}{1+\exp(-\betatnothat - \betaTprimehat \XmT)},
\end{align}
where $\betatnothat$ and ${\betaThat \coloneqq (\hat{\boldsymbol{\beta}}{}^0, \hat{\boldsymbol{\beta}}{}^{\delta}, \dots, \hat{\boldsymbol{\beta}}{}^{p\delta})^\top}$ are estimated regression coefficients and ${\XmT \coloneqq (\taum{0}, \taum{\delta}, \dots, \taum{p\delta})^\top}$ denotes the combined vector of travel times. Here, $\sigma(z) = \frac{1}{1+e^{-z}}$ is the logistic function that constraints the linear output $\betatnothat + \betaTprimehat \XmT$ to $(0,1)$.\footnote{This logistic hypothesis is commonly in discrete choice modeling \cite{benakiva1985}. We apply it in this work towards aggregate modal split prediction.} Hence, $\yihat$ can alternatively be interpreted as the probability that a traveler in time interval $t$ on day $i$ chooses to drive rather than take public transit.

We can then estimate $\betatnothat, \betaThat$ by maximizing the log-likelihood,
\begin{align*}
    \ell(\betatnot, \betat \mid \boldsymbol{\mathrm{q}}^1, \boldsymbol{\mathrm{q}}^0, \XT) = & \sum_{i=1}^m \Big[\mathrm{q}_{i}^1 \log\left(\sigma(\betatnot + \betaTprime \XmT)\right) + \\
    & + \mathrm{q}_i^0\log\left(1-\sigma(\betatnot + \betaTprime \XmT)\right)\Big], \nonumber
\end{align*}
where $m$ is the total number of days (i.e. observations) in our dataset. This formulation is equivalent to estimating a logistic regression with $\sum_{i=1}^m \mathrm{q}_i^1 + \mathrm{q}_i^0$ total observations of mode choices such that day $i$ contributes $\mathrm{q}_{i}^1$ driving and $\mathrm{q}_{i}^0$ transit observations, all corresponding to the same travel time vector $\XmT$.

\subsection{Estimation Methods}\label{sec:method}
We estimate the constant ${\betatnothat \in \mathbb{R}}$ and coefficients ${\betaThat \in \mathbb{R}^{n(p+1)}}$ using the dataset ${\XT \in \mathbb{R}^{m \times n(p+1)}}$, which contains $m$ days of travel time vectors in time periods $t$, $(t-\delta), \dots, (t-p\delta)$. For each day $i \in \{1,\dots,m\}$, the input data for interval $t$ is the vector $\XmT$. In many practical settings, constructing a dataset with observations spanning longer durations (many months or years) is not only difficult but also unadvisable given the changing behavior of travelers and network operations (e.g. changes to tolling, transit schedules, road closures, etc.). For these reasons, it is typically the case that $n(p+1) \gg m$; in our case study, we have $n(p+1)=1106$ and $m=130$. As such, estimating the regression function in \eqref{eq:logistic} suffers from multicollinearity and overfitting. To resolve challenges associated with high dimensionality, we use regularized logistic regression methods for shrinkage and variable selection \cite{friedman2010}.

For all regularization methods, we standardize the dataset $\XT$ to have a mean of $\boldsymbol{0}$ and standard deviation of $\boldsymbol{1}$ before estimation so that regularization is applied evenly across all variables \cite{hastie2001}.

\textit{1) Forward-stepwise selection} is a greedy algorithm for selecting a subset of variables with which we estimate a logistic regression \cite{hastie2001}. We initialize the algorithm with a constant model and iteratively incorporate the variable that, in combination with all previously selected variables, yields the lowest cross-validation error. This process continues until (i) there is no remaining variable such that adding that variable achieves a lower cross-validation error than that of the previous iteration, or (ii) adding a variable results in a rank-deficient data matrix with more variables than observations.

Forward-stepwise selection is computationally costly as it can require up to $\sum_{i=0}^{m-1} [n(p+1) - i]$ iterations of cross-validation.

\textit{2) Lasso regression} controls variance by imposing a penalty on the $\ell^1$ norm, or sum of absolute values, of the regression coefficients to yield a sparse coefficient estimate in which some coefficients are exactly zero \cite{hastie2001}. The lasso estimator maximizes the objective function,
\begin{align*}
    \max_{\betatnot, \betaT} \Big[ \ell(\betatnot, \betat \mid \boldsymbol{\mathrm{q}}^1, \boldsymbol{\mathrm{q}}^0, \XT) - \lambda\|\betat\|_1 \Big],
\end{align*}
where the hyperparameter $\lambda > 0$ governs the penalty imposed on the $\ell^1$ norm of the coefficient estimates. For higher values of $\lambda$, lasso automatically selects a smaller subset of variables and sets the coefficients of the remaining variables to zero.

\textit{3) Ridge regression} controls variance by imposing a penalty on the $\ell^2$ norm, or sum of squared values, of the regression coefficients to yield coefficients estimates that are shrunk towards zero \cite{hastie2001}. The ridge estimator maximizes the objective function,
\begin{align*}
    \max_{\betatnot, \betaT} \Big[ \ell(\betatnot, \betat \mid \boldsymbol{\mathrm{q}}^1, \boldsymbol{\mathrm{q}}^0, \XT) - \lambda\|\betat\|_2^2 \Big],
\end{align*}
where the hyperparameter $\lambda > 0$ governs the penalty imposed on the $\ell^2$ norm of the coefficient estimates. Higher values of $\lambda$ yield more shrunken coefficient estimates.\footnote{Other methods for regularization and variable selection are also commonly used, such as elastic net and principal components regression. We do not include them here as they did not perform substantially better in our case study than the listed methods in our empirical study.}

\subsection{Model Selection and Evaluation}\label{sec:crossval}
We select the hyperparameter of each method (i.e. $\lambda$ for ridge and lasso regression) and the set of predictors in forward-stepwise selection by minimizing the average out-of-sample root mean squared error (RMSE) in a $K$-fold cross-validation. We randomly split the dataset into $K$ folds $I_1, \dots, I_{K}$. For each fold $I_f$ where $f=1, \dots, K$, we estimate $\betatnothat, \betaThat$ using data in the remaining folds $\{1,\dots,m\} \setminus I_f$. We then calculate the out-of-sample root mean squared error (RMSE) for fold $I_f$ as
\begin{align*}
    \mathrm{RMSE}_f &= \sqrt{\frac{1}{|I_f|} \sum_{i \in I_{f}} \left(\mathrm{y}_i - \sigma(\betatnothat + \betaTprimehat \XmT)\right)^2}.
\end{align*}
The average cross-validation RMSE is an absolute measure of error that describes the standard deviation of out-of-sample prediction errors. Note that this metric has the same units as the driving fraction, i.e. the fraction of travelers choosing the freeway network. Practically, however, the driving fraction shows greater variation, and consequently decreased predictability, in some hours compared to others (see. Fig. \ref{fig:driving_fraction}). To compare prediction error across hours, we consider the out-of-sample $R^2$, which we compute for each fold $I_f$ as follows:
\begin{align*}
    R^2_f = 1-\frac{\sum_{i \in I_{f}} \left(\mathrm{y}_i - \sigma(\betatnothat + \betaTprimehat \XmT)\right)^2}{\sum_{i \in I_f} \left(\mathrm{y}_i - \overline{\mathrm{y}} \right)^2}.
\end{align*}
Here, $\overline{\mathrm{y}}$ is the sample mean of $\mathrm{y}_i$ for $i\in I_f$. The out-of-sample $R^2$ normalizes the mean squared error by the sample variance of the driving fraction for each hour, which allows comparison of performance across hours. Unlike the in-sample $R^2$, the out-of-sample $R^2$ is not lower bounded by $0$ and instead ranges from $-\infty$ to $1$ with values closer to $1$ indicating higher out-of-sample accuracy. For any given hyperparameter(s), we iterate this process over all folds and compute the average cross-validation RMSE and out-of-sample $R^2$ as
\begin{align*}
    \overline{\mathrm{RMSE}} = \frac{1}{K}\sum_{f=1}^K \mathrm{RMSE}_f, \hspace{5mm} \overline{R^2} = \frac{1}{K}\sum_{f=1}^K R^2_f.
\end{align*}
\section{Interpretation Framework}\label{sec:interpretation}
\subsection{Pre-specified variable selection}\label{sec:pre-specified}
We introduce two approaches to variable selection as benchmarks to evaluate the regularized models described in Sec. \ref{sec:method}. In particular, we compute a metric, either betweenness centrality or average speed, for each segment and regress on a subset of segments that fall within a given range for each metric. Selecting these subsets and comparing their estimated coefficients to those from regularized approaches allows us to identify how topological (i.e. betweenness) and temporally varying demand-based (i.e. average speed) segment characteristics are selected in different time intervals.

\subsubsection{Selection based on betweenness centrality} We define a betweenness centrality metric for each segment as the number of routes -- sequences of segments -- that connect an origin in region A to a destination in region B and pass through the given segment. From a topological viewpoint, a segment with higher betweenness centrality affects the travel time of more routes that pass through it. The key hypothesis here is that changes in travel time on segments with a high betweenness centrality will likely have a larger impact on the aggregate driving fraction relative to those with a lower centrality. Consequently, in building a logistic regression, one can select the travel time variables of segments with betweenness centralities above a certain threshold $b$.

\subsubsection{Selection based on average speed} We compute the average speed of each segment as the ratio between the segment length and travel time. Segments with low average speeds are considered to be more congested, and consequently changes in travel time on these segments are likely to have a greater impact on the driving fraction $y$. Thus, an alternative to variable selection based on betweenness centrality is to select the travel time variables of segments with average speeds below a certain threshold $v$.

For both the betweenness and speed selection methods, we also select the travel time variables of segments observed in time $(t-k_b \delta)$ and $(t-k_v \delta)$, respectively. Here, the time lags $k_b$ and $k_v$ are also hyperparameters.

We estimate the logistic regression coefficients of the travel times on the segments selected based on betweenness centrality or average speed by maximizing the log-likelihood, ${\ell(\betatnot, \betat \mid \boldsymbol{\mathrm{q}}^1, \boldsymbol{\mathrm{q}}^0, \XT)}$, where $\XT$ contains travel times for the selected segments.

\subsection{Degrees of freedom}\label{sec:degrees}
For any estimated model, we report the effective degrees of freedom, $\mathrm{df} < n(p+1)$, which can be viewed as a measure of model complexity. For methods that yield sparse coefficient estimates, $\mathrm{df}$ is the number of non-zero coefficients. For ridge regression, $\mathrm{df}$ can be computed as $\mathrm{Trace}(\XT(\XTprime \XT +\lambda \mathbf{I})^{-1}\XTprime)$, where $\mathbf{I}$ is the identity matrix. Note that $\mathrm{df}$ is closely related to the regularization hyperparameter of a given model as greater regularization yields lower values of $\mathrm{df}$.

This quantity provides a behavioral interpretation -- higher values of $\mathrm{df}$ suggest more travel time information is incorporated in the model prediction, i.e. the aggregate driving fraction can be better predicted with travel time information from a larger set of network segments. However, $\mathrm{df}$ is not a measure of responsiveness (which we discuss in the following subsection) as $\mathrm{df}$ is the result of estimating a predictive model with a limited number of observations and many correlated predictor variables.

\subsection{Behavioral implications}\label{subsec:correlation}
From a behavioral viewpoint, one can expect that the changes in travel times of certain ``critical'' segments would impact travelers' preferences for the routes on which these segments lie. Consequently, for any given hour, such segments can have a significant impact on the modal split. We argue that our modeling approach is useful for identifying critical segments in each hour and helps us study how these segments impact the travelers' aggregate responsiveness to travel time changes. In Sec. \ref{sec:empirical}, we show that this behavioral interpretation provides new insights about high \textit{variability} and low \textit{responsiveness} of the modal split (and hence difficulty in its prediction) for certain hours of the day. We now formally introduce these quantities below.

\subsubsection{Variability}
For any time interval $t$, we quantify the inherent variability of the mode share $\sigma^2$ in time $t$ by the variance of the driving fraction, i.e.
\begin{align*}
    \sigma^2 &= \frac{1}{m}\sum_{i=1}^m (\mathrm{y}_i - \overline{\mathrm{y}})^2.
\end{align*}
Time intervals $t$ with higher values of $\sigma^2$ correspond to more difficult prediction tasks.

\subsubsection{Responsiveness}
For any time interval $t$, any freeway segment $j$, and any time lag $k$, we denote the correlation ratio between the travel time of segment $j$ in time $(t-k\delta)$ with the driving fraction $\YT$ as $\mathrm{Corr}(\YT, \tau_j^{k\delta}) \in (0,1)$, which is given by
\begin{align*}
    &\mathrm{Corr}(\YT, \tau_j^{k\delta}) = \\
    &=\frac{\sum_{i=1}^m (\mathrm{y}_i - \overline{\mathrm{y}})(\tau_{i,j}^{k\delta} - \overline{\tau}{}_{j}^{k\delta})}{\sqrt{\sum_{i=1}^m {(\mathrm{y}_i - \overline{\mathrm{y}})}^2}\sqrt{{(\tau_{i,j}^{k\delta} - \overline{\tau}{}_{j}^{k\delta})}^2}},
\end{align*}
where $\overline{\tau}{}_{j}^{k\delta} = \frac{1}{m} \sum_{i=1}^m \tau_{i,j}^{k\delta}$. We quantify the responsiveness of mode share to changes in travel time by 
\begin{align*}
    \rho = \frac{1}{n(p+1)} \sum_{k=0}^p \sum_{j=1}^n \mathrm{Corr}(\YT, \tau_j^{k\delta}),
\end{align*}
or the correlation between the driving fraction $\YT$ and the travel times $\tau_j^{k\delta}$ averaged over all segments $j$ and time lags $k$.

The value of the average correlation ratio $\rho$ is in $(-1, 1)$. Negative values suggest that high travel times of freeway segments reduce the driving fraction, while positive values suggest that high driving times increase the driving fraction. Moreover, the average correlation value being closer to $-1$ indicates that the driving fraction is more responsive to the driving times of the segments.
\section{Empirical Study in San Francisco Bay Area}\label{sec:empirical}
\begin{figure}[htp]
    \centering
    \includegraphics[width=\columnwidth]{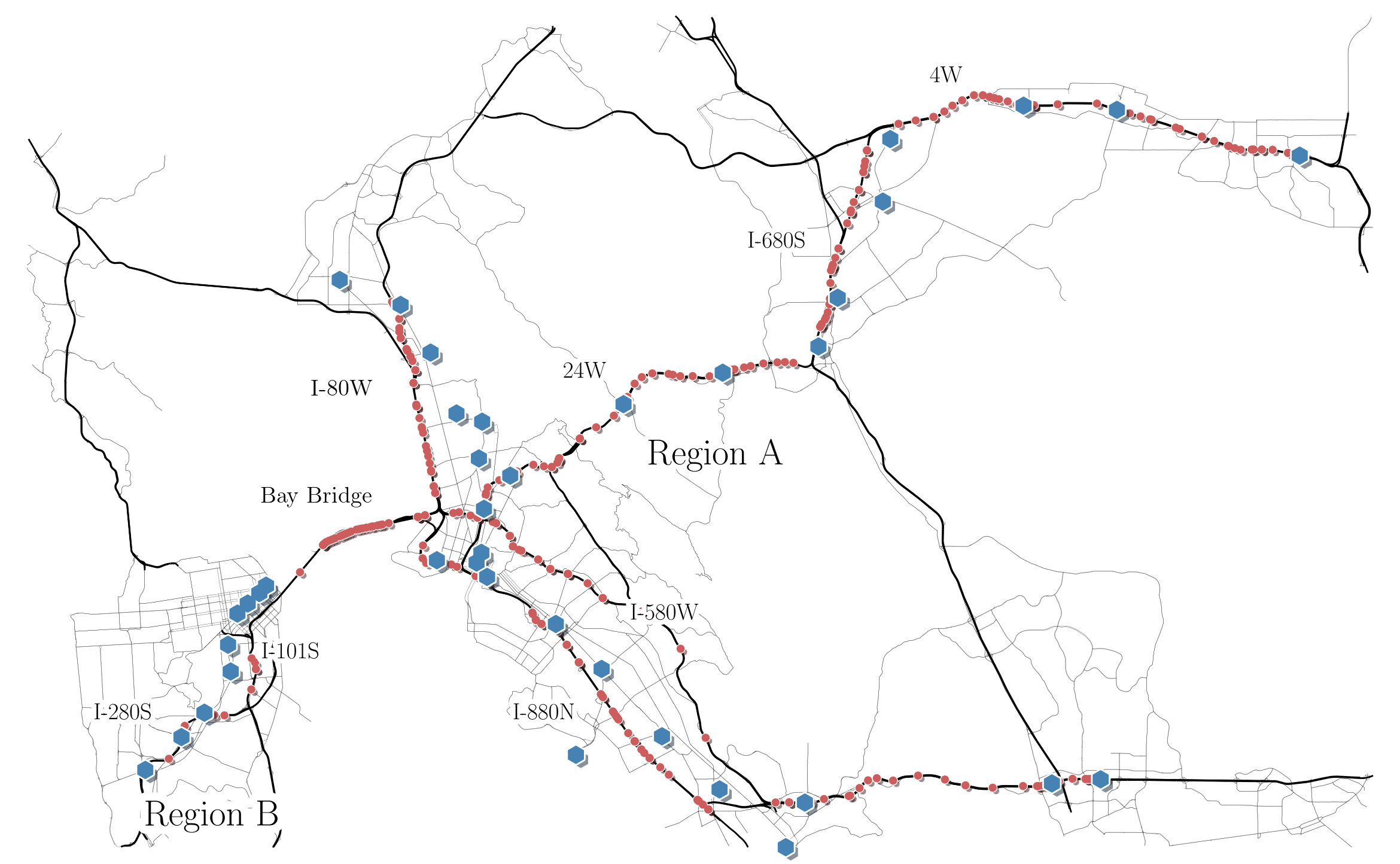}
    \caption{PeMS loop detectors (red) and BART stations (blue) in the San Francisco Bay Area.}
    \label{fig:map}
\end{figure}

The goal of our case study is to predict the fraction of driving flows during each one-hour time interval from 5am to 10pm in the SF Bay Area. Recall that our multi-modal network is comprised of the freeway network and the BART network. The origin region A includes the East Bay, and the destination region B includes San Francisco proper and Daly City. The two regions are connected by the San Francisco - Oakland Bay Bridge freeway and the transbay BART tube. The shortest driving route from any origin in A to any destination in B passes through the Bay Bridge, and similarly any BART route that connects stations in the two areas passes through the transbay tube.\footnote{The boundaries of A and B are defined such that any freeway route from A to B that passes other bridges requires significant length of detour. All BART lines that cross the bay go through the transbay tube. We do not consider BART trips that end in the West Bay south of Daly City as the shortest freeway route to these destinations from some origins in A might be through the San Mateo - Hayward Bridge.} The network includes 158 freeway segments (determined by neighboring on-ramps and off-ramps) and 38 BART segments (determined by adjacent stations). Regions A and B as well as relevant freeways, loop detectors, and BART stations are depicted in Fig. \ref{fig:map}.

\subsection{Data}\label{subsec:data}
We use traffic flow data reported by the Caltrans Performance Measurement Systems (PeMS) \cite{pems} and transit ridership data reported by the BART authority on workdays between February and September of 2019. We identify and remove any days in which severe incidents have been reported in the California Highway Patrol (CHP) incidents report dataset to have caused lane closures in the two areas as well as any days in which BART operations are severely disrupted.\footnote{This is because travelers may not have complete information of the incidents, and thus the driving and transit flows induced by their mode choices during incident hours may not fully account for the non-recurrent delays caused by these incidents.} In total, we use data from $m=130$ workdays. 

Next, we describe how we measure the hourly driving flow, BART ridership, and average travel time of segments. 
\begin{enumerate}
    \item \textit{Driving flow.} The PeMS data includes the hourly driving flow (i.e. the number of passing vehicles in each hour) recorded by loop detector sensors that are embedded under the freeways. Since any shortest route that connects regions A and B passes over the bridge, the total driving flow from A to B is the flow recorded by the loop detectors on the bridge.
    \item \textit{BART ridership.} The tap-in/tap-out systems at BART stations record the number of completed trips between each pair of stations in each hour. We compute the total ridership from A to B by summing over all station pairs with origins in A and destinations in B. 
    \item \textit{Average travel time of segments.} The PeMS dataset records the average speed of vehicles that pass over each loop detector in every 5-minute interval. We compute the average driving time of each segment as the ratio between the segment length and the average speed recorded by the loop detectors located in that segment.\footnote{The PeMS dataset also reports performance metrics for each detector -- the percentage of traffic counts that is not imputed. We use the reported detector performance to identify and delete data points collected by loop detectors that have less than 80\% non-imputed traffic counts. For segments with no working detectors, we estimate the segment speed as the average speed reported by the closest detectors immediately upstream and downstream of the segment.} The operating time for each BART segment is not recorded by the BART authority, and thus cannot be used in our prediction.\footnote{The variance of BART travel time is much less than that of driving. According to the ``Customer On-Time Performance" report provided by BART, over 90\% of BART trips were made on time for most days in 2018.}
\end{enumerate} 

\begin{figure}[htp]
    \centering
    \begin{subfigure}{\columnwidth}
        \includegraphics[width=\columnwidth]{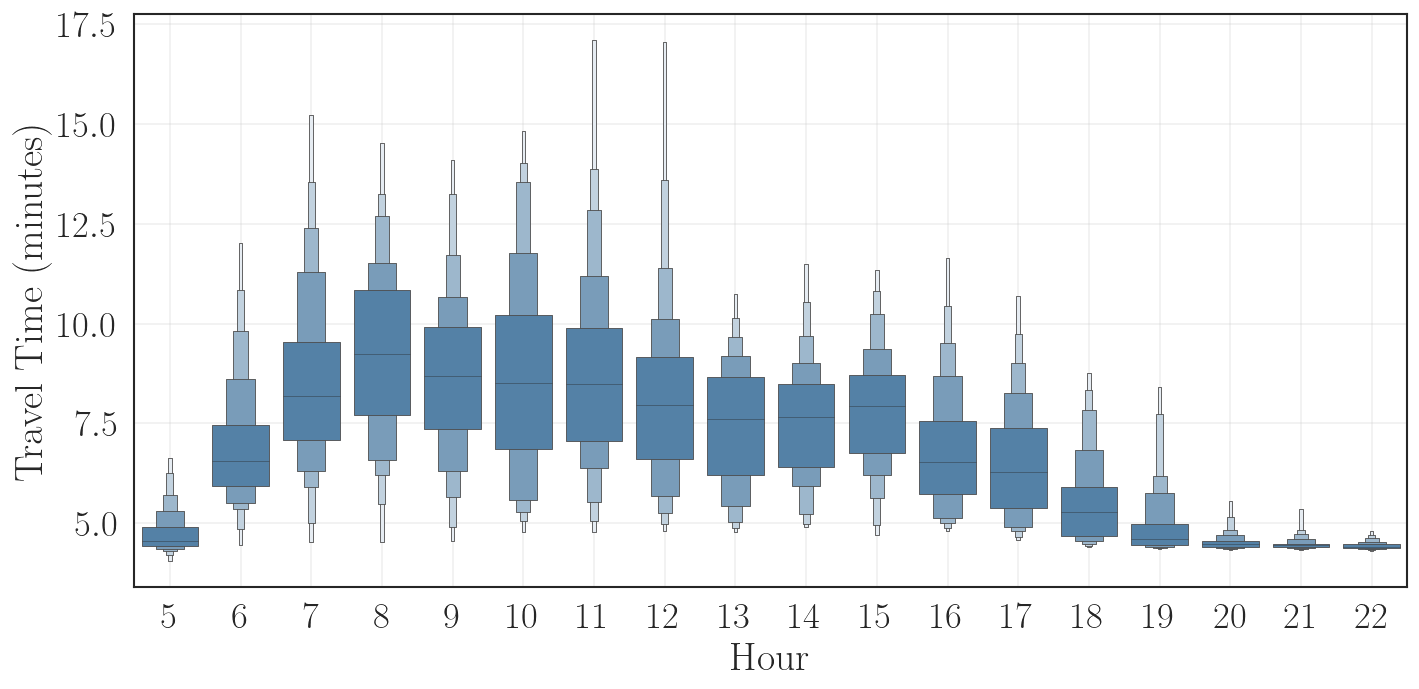}
        \caption{Driving travel times on I-80W peak in morning rush.}
        \label{fig:driving_time}
    \end{subfigure}
    \begin{subfigure}{\columnwidth}
        \includegraphics[width=\columnwidth]{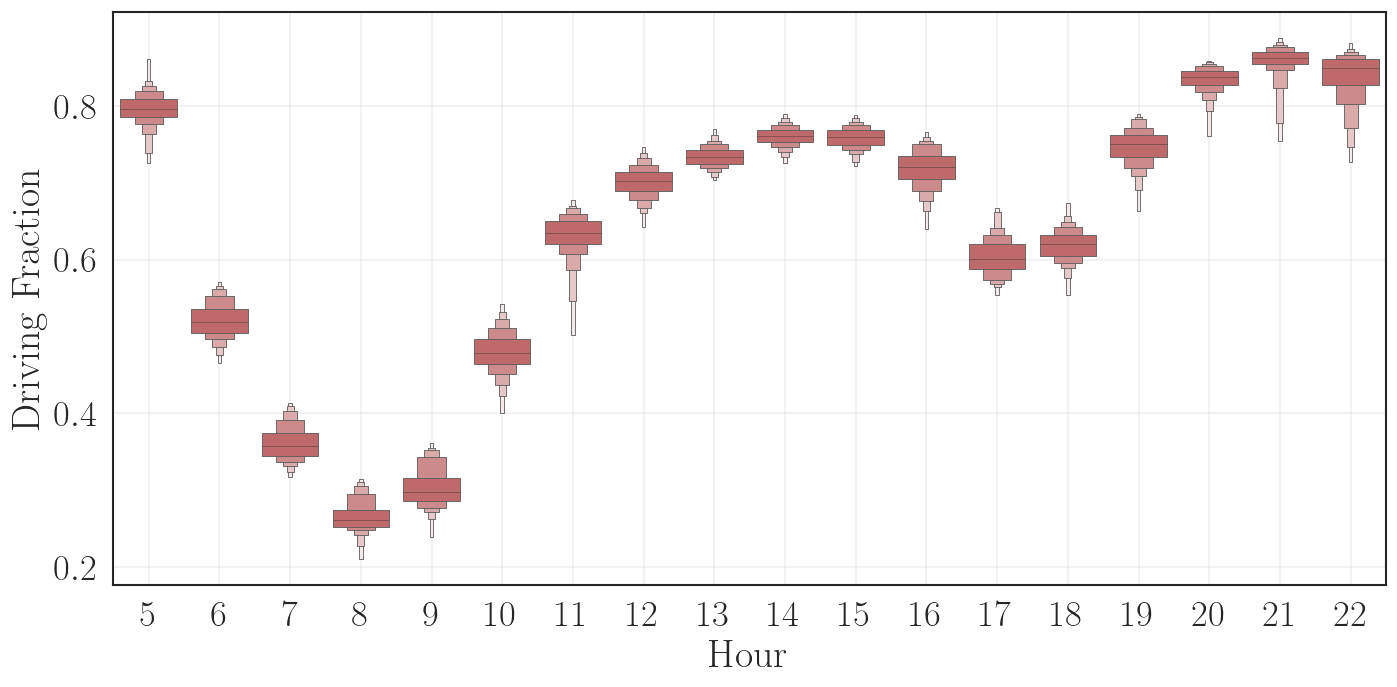}
        \caption{Driving fraction is lowest in morning rush hour.}
        \label{fig:driving_fraction}
    \end{subfigure}
    \caption{Travel times and modal split vary over time.}
    \label{fig:network_statistics}
\end{figure}

Fig. \ref{fig:driving_time} shows the variability of driving travel times from Richmond to the Bay Bridge along the north branch of our network (i.e. I-80W). The highest driving time occurs during morning rush hour and is three times the lowest driving time, which occurs at night. We also observe that the confidence interval of the driving time in each hour is quite wide, which suggests that travelers experience high variability in day-to-day driving times. Furthermore, Fig. \ref{fig:driving_fraction} demonstrates that the driving fraction is lowest during morning rush hour and highest at night. We also observe within-day variation in the driving fraction. This suggests that the driving fraction induced by travelers' choices between driving and taking public transit might be sensitive to the changes of driving time in the network over the course of the day.

\subsection{Results and discussion}\label{subsec:results}

\begin{figure}[htp]
    \centering
    \begin{subfigure}{0.414\textwidth}
        \includegraphics[width=\textwidth]{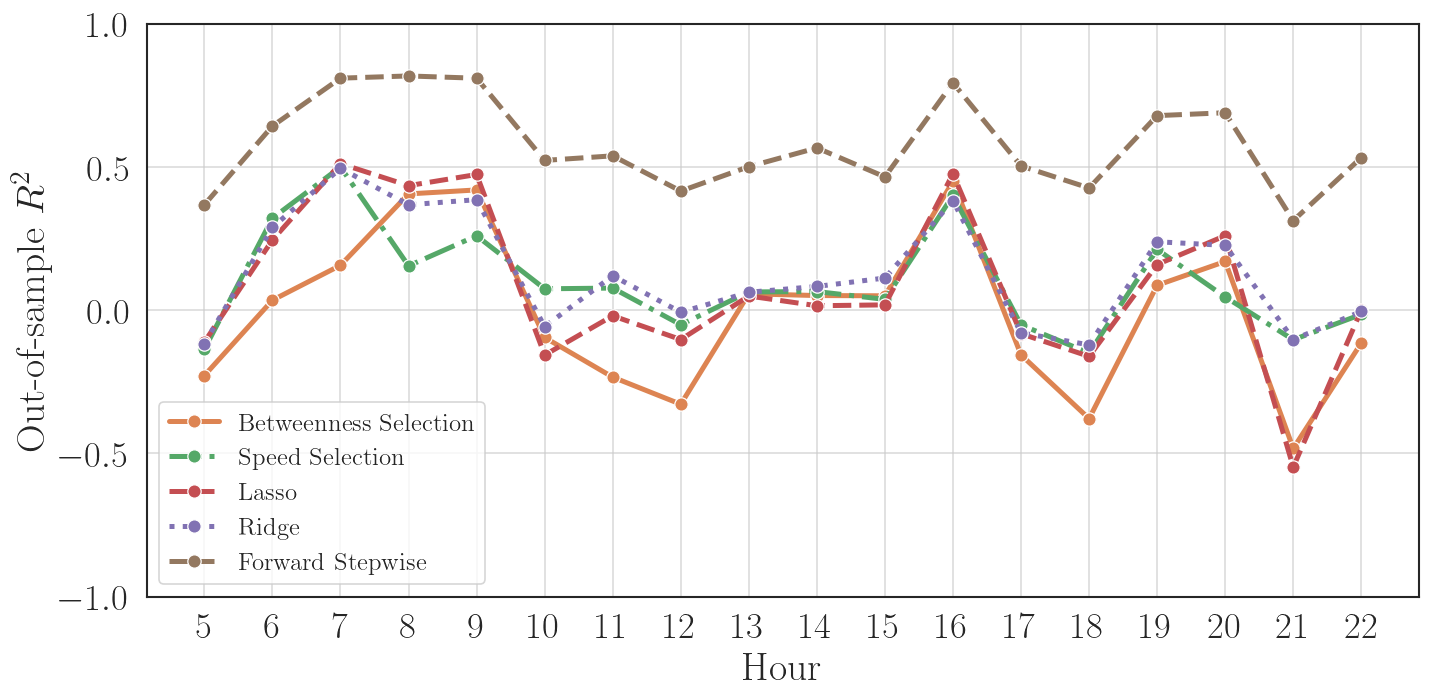}
        \caption{Out-of-sample $R^2$ for each method by hour.}
        \label{fig:rsquared}
    \end{subfigure}
    \par\bigskip
    \begin{subfigure}{0.414\textwidth}
        \centering
        \includegraphics[width=\textwidth]{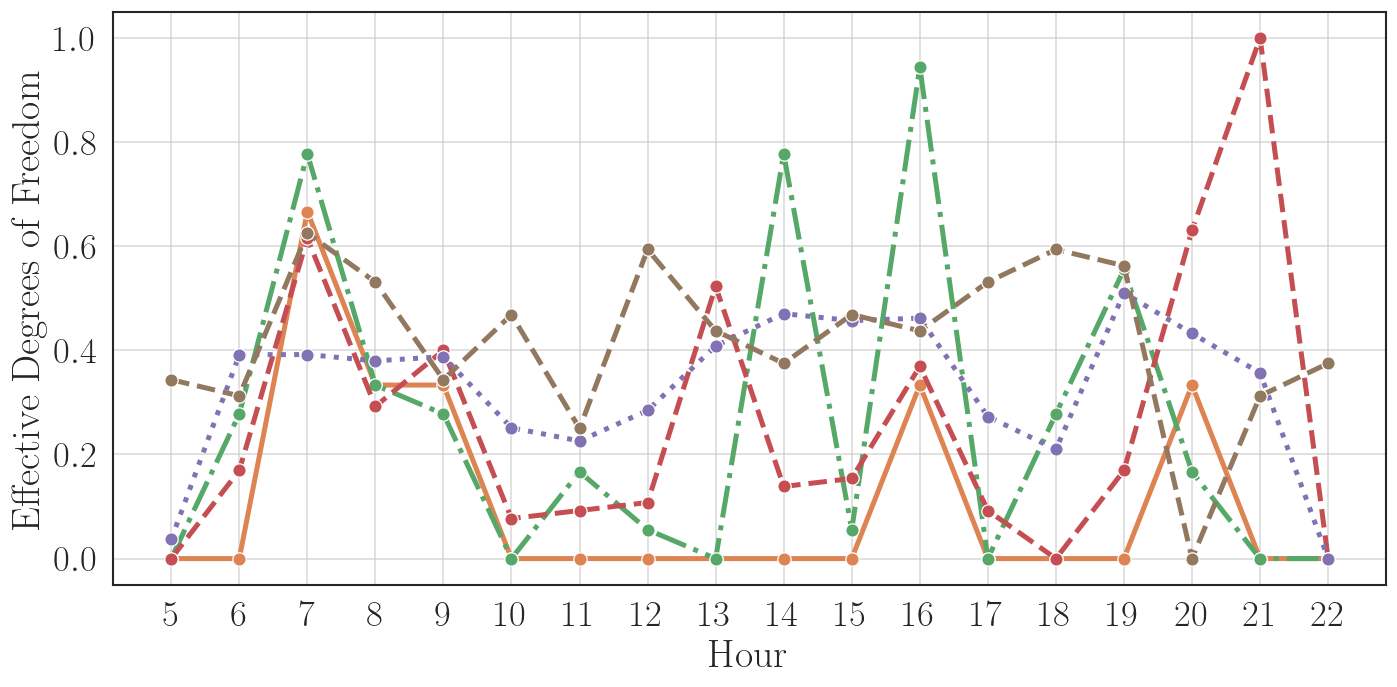}
        \caption{Effective degrees of freedom for each method by hour.}
        \label{fig:degrees}
    \end{subfigure}
    \caption{All methods demonstrate high $R^2$ in the morning. Model complexity peaks in the morning, mid-afternoon, and evening for all methods except forward-stepwise selection.}
\end{figure}

The results from our case study are summarized in the following figures: (i) Fig. \ref{fig:rmse_results} compares the predictive accuracy of various models by reporting the final 5-fold cross validation RMSE for each variable selection method (see Sec. \ref{sec:method} and \ref{sec:pre-specified}) in each hour. For the sake of comparison, we also include results from applying a random forest approach -- a popular nonparametric ML method \cite{hastie2001}. (ii) Relative accuracy is reported in  Fig. \ref{fig:rsquared} as the average cross-validation out-of-sample $R^2$ results (see \ref{sec:crossval}) for each method in each hour. (iii) Model complexity is reported in Fig. \ref{fig:degrees} as the effective degrees of freedom (see Sec. \ref{sec:degrees}) selected by our 5-fold cross-validation for each method and each hour.

\begin{figure*}[htp]
    \centering
    \includegraphics[width=0.8\textwidth]{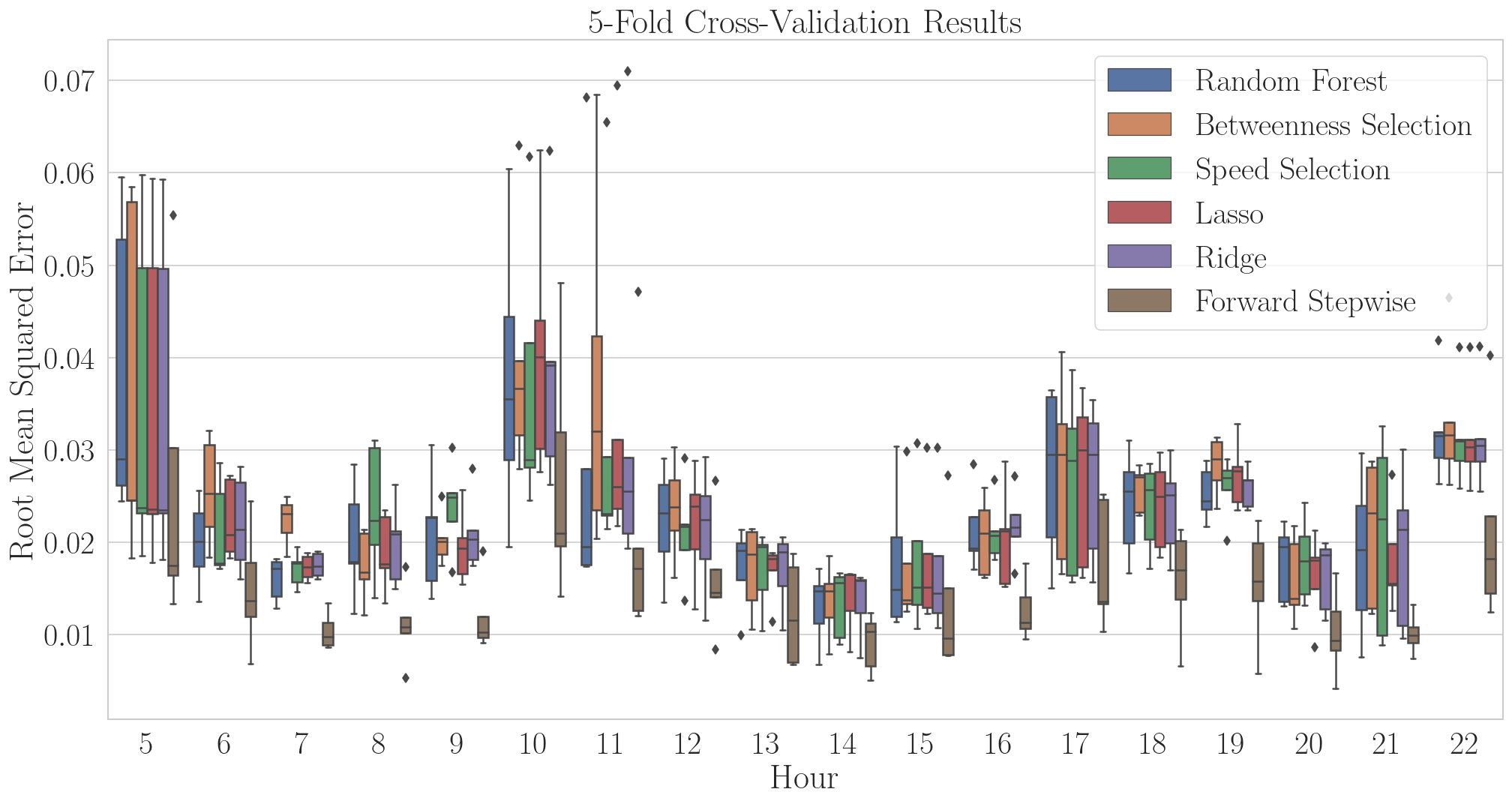}
    \caption{Horizontal lines inside boxes represent the median cross-validation RMSE values. Boxes extend from the first (Q1) to the third quartile (Q3) while the whiskers extend beyond the box by 1.5 times the interquartile range (Q3-Q1). Observations beyond the whiskers are considered outliers.}
    \label{fig:rmse_results}
\end{figure*}

\begin{figure}[htp]
    \centering
    \begin{subfigure}{0.45\textwidth}
        \includegraphics[width=\textwidth]{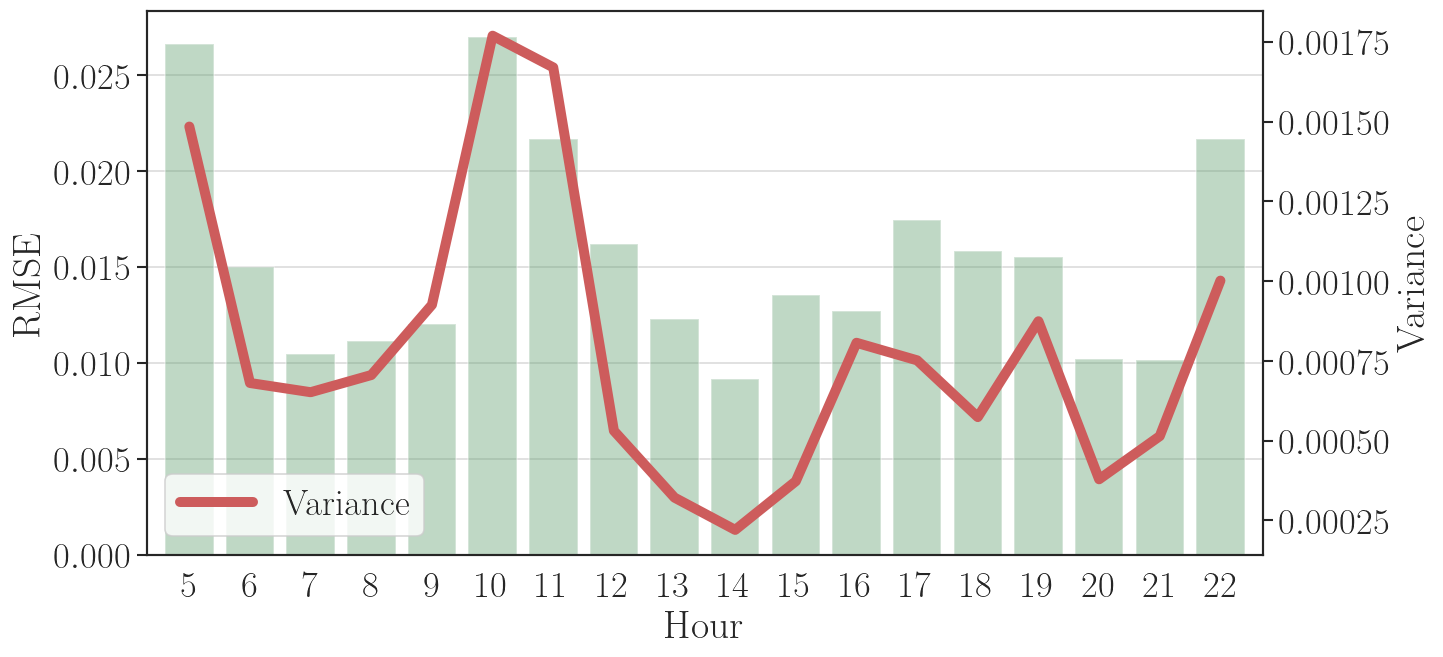}
        \caption{Forward-stepwise selection RMSE vs. variability.}
        \label{fig:variance}
    \end{subfigure}
    \par\bigskip
    \begin{subfigure}{0.45\textwidth}
        \includegraphics[width=\textwidth]{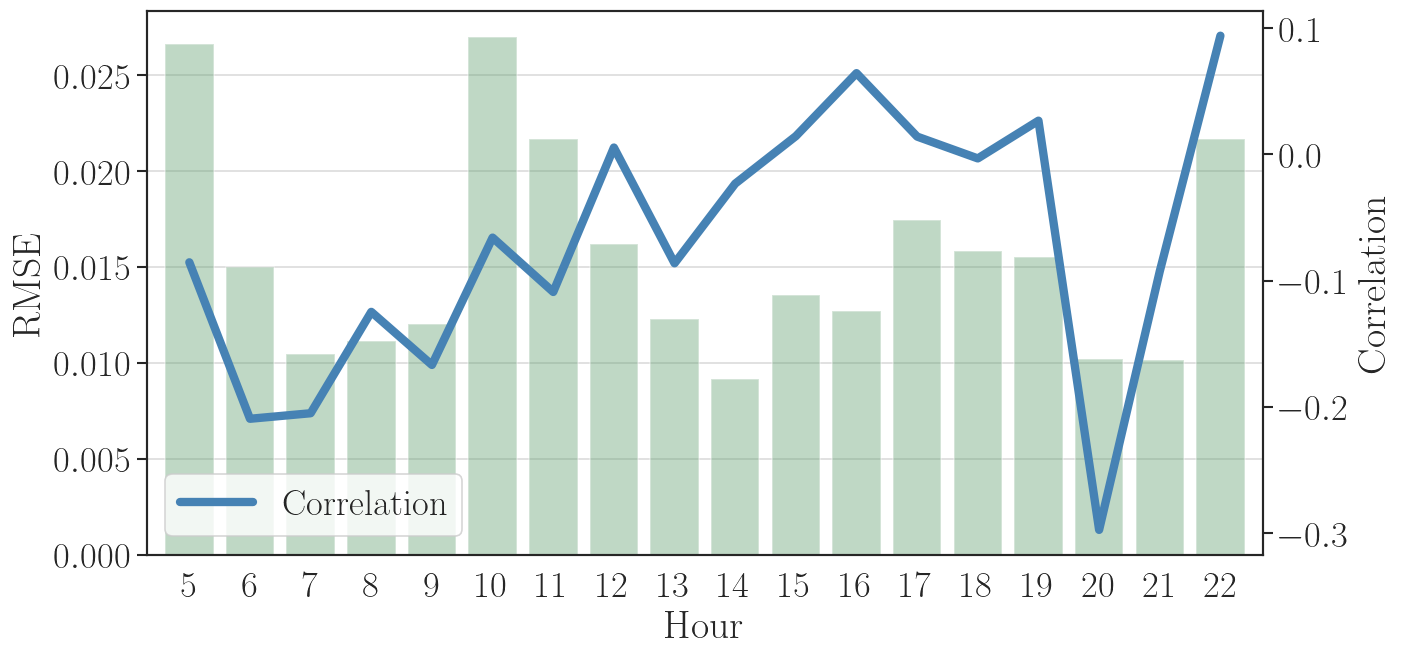}
        \caption{Forward-stepwise selection RMSE vs. responsiveness.}
        \label{fig:correlation}
    \end{subfigure}
    \caption{Varying predictive performance across hours can be partially explained by modal split variability and responsiveness.}
\end{figure}

Based on these results, we can make the following observations: 
\subsubsection{Overall performance} All methods provide accurate prediction of driving fraction with a median RMSE below 0.03 across all hours except for the hours of 10am, 11am, and 10pm.
\subsubsection{Comparison across hours} Our models achieve their highest predictive performance during morning rush hour (6am--9am) and early afternoon (12pm--4pm). The median RMSE and variance of RMSE across folds in these hours are both lower than in other hours.

One reason for superior predictive performance in morning rush hour is that travelers are more likely to respond to changes in driving time within these hours. Indeed, from Fig. \ref{fig:correlation}, we can see that 6am--9am are hours in which the driving fraction exhibits higher responsiveness (i.e. more negative correlation) towards the changes in driving time. Additionally, we can see from Fig. \ref{fig:rsquared} that these hours achieve a higher out-of-sample $R^2$ and also select for high model complexity (i.e. the driving time of more segments are selected in the regression) in all methods except forward-stepwise selection (Fig. \ref{fig:degrees}). On the other hand, high prediction accuracy in the early afternoon (12pm--4pm) and evening (8pm--9pm) is a result of especially low variability rather than high responsiveness. Finally, we observe that, despite exhibiting moderate responsiveness, prediction accuracy in 10am--11am suffers as a result of high variability. In summary, observations suggest that prediction accuracy in a given hour is restricted by moderate variability unless high responsiveness is also exhibited.

\subsubsection{Comparison across methods}
In Fig. \ref{fig:rmse_results}, we observe that the forward-stepwise selection approach achieves a lower RMSE than all other methods. This is likely due to the fact that forward-stepwise selection chooses from a larger set of combinations of predictors.

We also observe that the regularized methods generally achieve a lower RMSE than the two pre-specified variable selection methods (see Sec. \ref{sec:pre-specified}). This is due to the fact that these regularized methods address model overfitting and multicollinearity in a more flexible manner than variable selection methods that are purely based on betweenness centrality or average speed. However, the performance of either pre-specified variable selection method relative to the other varies across hours. In 8am--9am, selection based on betweenness centrality outperforms selection based on average speed, which coincides with the hours of highest travel times. In almost all other hours, however, selection based on average speed outperforms selection based on betweenness centrality. This might be due to the fact that the highway network is already largely in a congested regime in 8am--9am, and during this time incorporating segments that are more congested is likely less beneficial for prediction than incorporating more central segments. In fact, the heatmaps of model coefficients presented in Sec. \ref{subsec:heatmaps} demonstrate that regularized methods are able to capture both congestion-based and spatial characteristics when assigning coefficients to segments.

\subsection{Spatial distribution of critical segments}\label{subsec:heatmaps}
\begin{figure}[b!]
     \centering
     \begin{subfigure}[b]{0.23\textwidth}
         \centering
         \includegraphics[width=\textwidth]{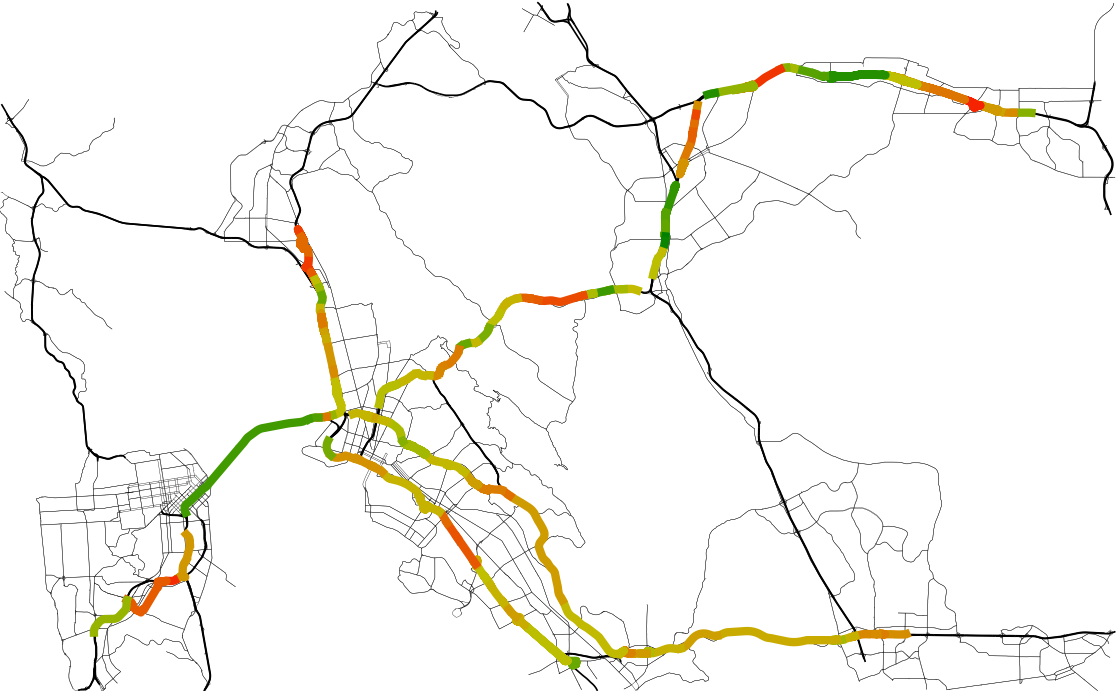}
         \caption{Ridge coefficients 7am}
     \end{subfigure}
     \hfill
     \begin{subfigure}[b]{0.23\textwidth}
         \centering
         \includegraphics[width=\textwidth]{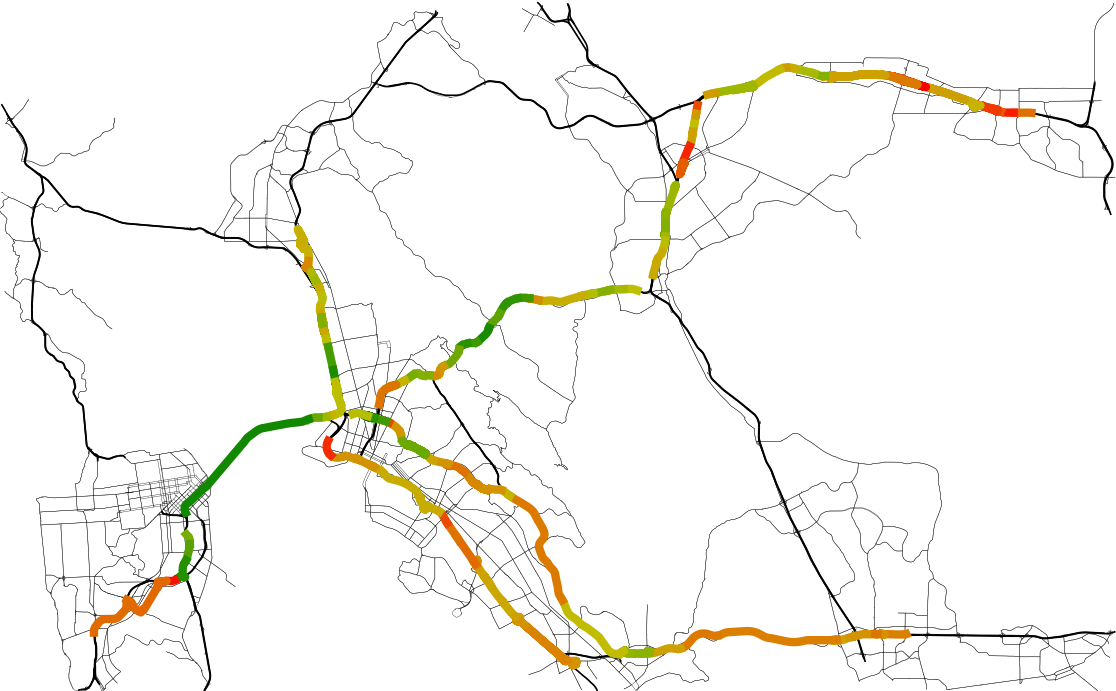}
         \caption{Ridge coefficients 9am}
     \end{subfigure}
     \par\medskip
     \centering
     \begin{subfigure}[b]{0.23\textwidth}
         \centering
         \includegraphics[width=\textwidth]{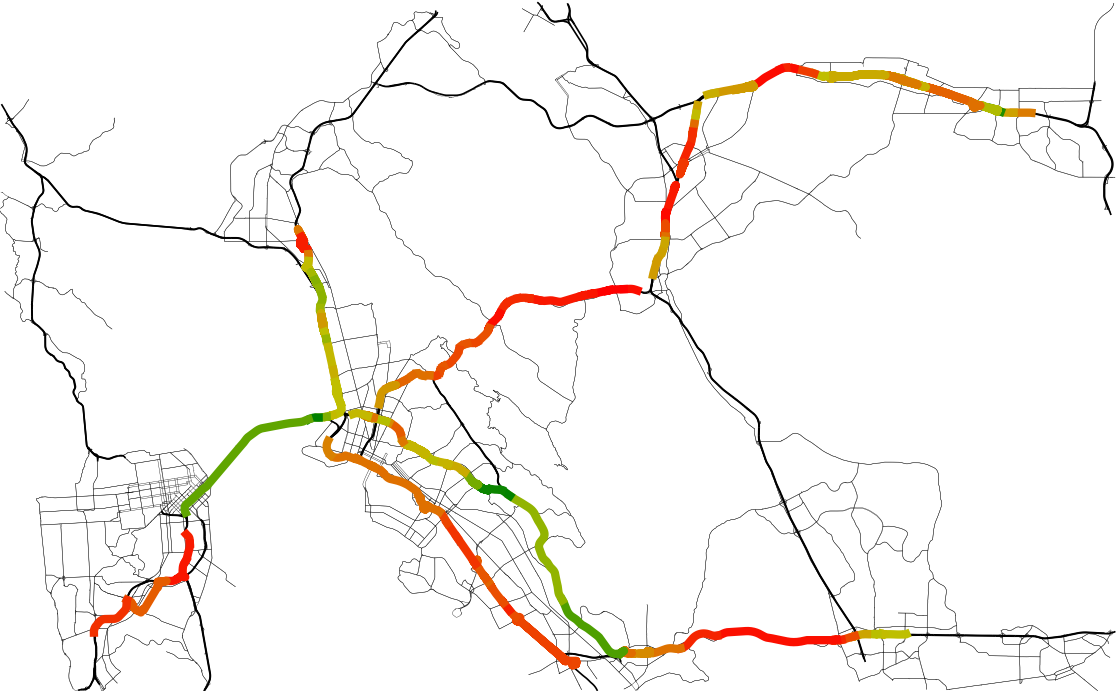}
         \caption{Ridge coefficients 11am}
     \end{subfigure}
     \hfill
     \begin{subfigure}[b]{0.23\textwidth}
         \centering
         \includegraphics[width=\textwidth]{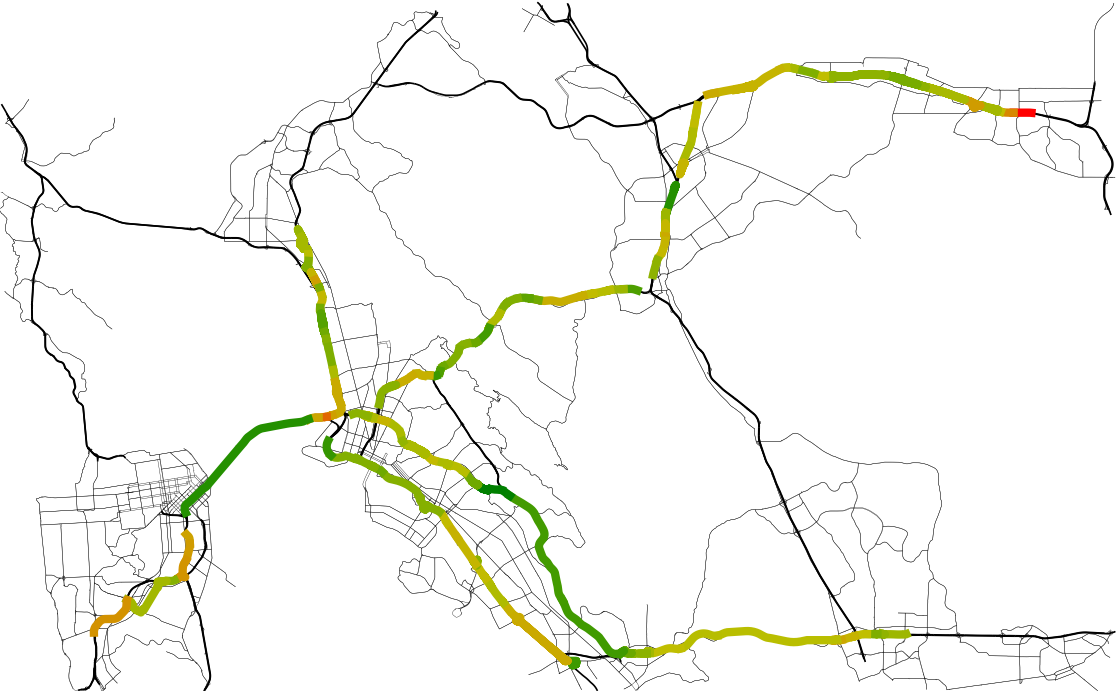}
         \caption{Ridge coefficients 1pm}
     \end{subfigure}
     \par\medskip
     \centering
     \begin{subfigure}[b]{0.23\textwidth}
         \centering
         \includegraphics[width=\textwidth]{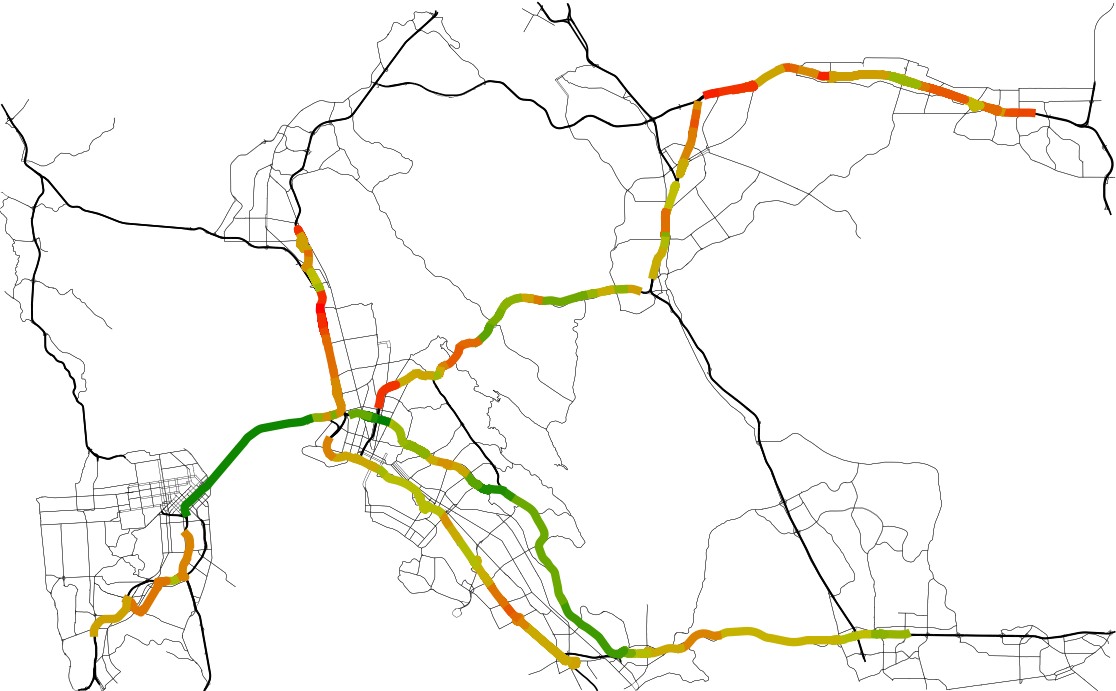}
         \caption{Ridge coefficients 3pm}
     \end{subfigure}
     \hfill
     \begin{subfigure}[b]{0.23\textwidth}
         \centering
         \includegraphics[width=\textwidth]{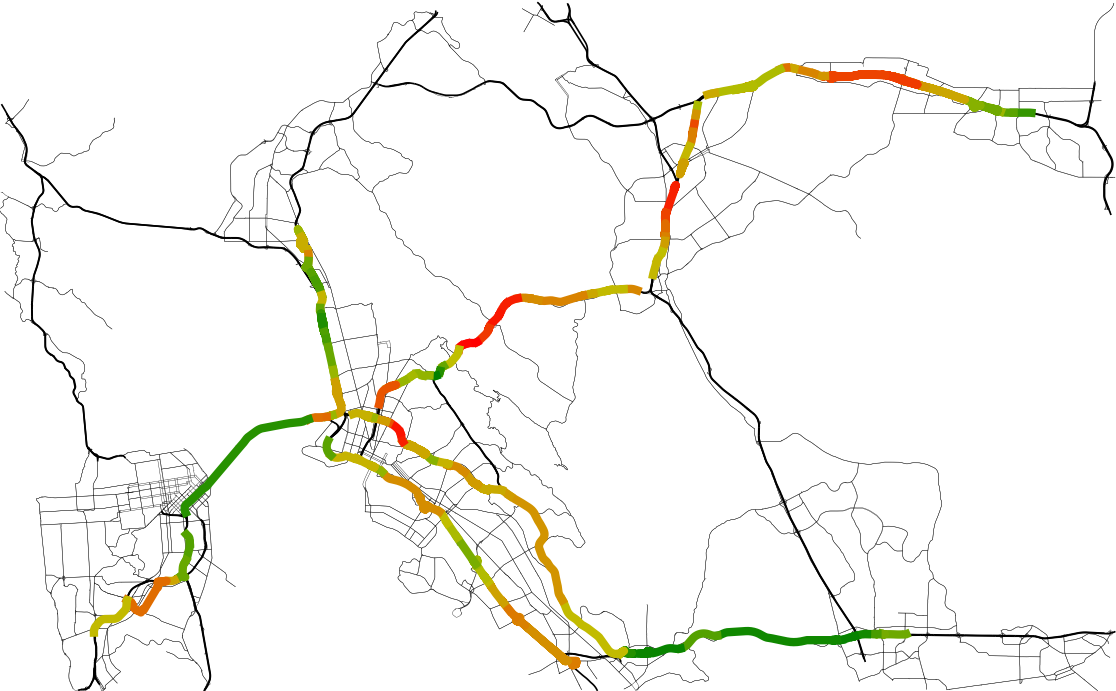}
         \caption{Ridge coefficients 5pm}
     \end{subfigure}
     \caption{Relative values of weights are colored on a spectrum from red to green, where {\color{BrickRed} red} represents lower weight (higher criticality) values and {\color{OliveGreen} green} represents higher weight (lower criticality) values. Weights are normalized in each hour to convey the geospatial distribution of weights so the weight of any one segment cannot be compared across hours.}
     \label{ridge}
\end{figure}

We compute the weight of each segment $j \in \{1,\dots,n\}$ as the sum of coefficients across all time lags, i.e. ${w_{j}=\sum_{k=0}^p \hat{\beta}{}_{j}^{k\delta}}$. We demonstrate the geospatial distribution of segment weights for ridge regression using heatmaps in Fig. \ref{ridge}.

Lower (more negative) weights indicate that an increase of driving time on the considered segment has a greater impact on reducing the overall driving fraction. Thus, segments marked in red are ones that have a high impact on the driving fraction and thus deemed ``critical'' for the prediction task. Practically, prior knowledge of these high impact segments can be useful to the transportation authority for effective management of driving and transit demand.

Coefficient heatmaps for all methods demonstrate a similar pattern of how the spatial distribution of high-impact segments changes across hours. In particular, we observe that segments on the northeast branch have a high impact on the modal split in most hours, while segments on the south and southeast branches demonstrate high criticality in 8am--11am.

Finally, we note that the set of high-impact segments identified by lasso, ridge, and forward-stepwise selection includes segments with high betweenness centrality (i.e. the Bay Bridge and surrounding segments) and segments with low speed (i.e. on the north and northeast branches). This suggests that the superior performance of these regularization methods in comparison to pre-specified selection methods can be explained by their flexibility in selecting for combinations of segments that are topologically central and/or highly congested. However, segments that are neither topologically central nor highly congested may also be selected; this may be explained by the sensitivity of demand along these segments to occasional congestion.

\section{Conclusion}

In this work, we propose an interpretable machine learning approach for predicting aggregate modal split for travelers in a multi-modal transportation network. We use a regularized logistic regression model, namely with $\ell^1$ norm and $\ell^2$ norm regularization as well as forward-stepwise variable selection. We discuss behavioral implications of our results by quantifying the inherent variability of the modal split as well as responsiveness of the modal split to changes in travel time. Finally, we apply our approach to predicting driving fraction in the SF Bay Area and discover that our models achieve their highest accuracy on predictions during: (1) morning rush hour (6am--9am) when travelers are most responsive to travel time, and (2) the early afternoon (12pm--4pm) when variability is lowest.

Visualizing our estimated model parameters as heatmaps allows us to identify critical regions in the freeway network at different hours. As such, our approach provides a class of accurate predictive models along with useful interpretations that may support transportation authorities in anticipating demand shifts in critical regions and making operational decisions accordingly.

Future work may focus on predicting aggregate modal split in a severe incident setting. Preliminary investigation suggests this task is more difficult than predicting aggregate modal split during nominal conditions. Accurate prediction in incident settings requires the construction of a larger incident event dataset as well as a different approach to analyzing the behavioral adjustments of travelers as they obtain incident information and reroute. Finally, designing resilient network operational decisions based on aggregate modal split predictions is another problem of significant interest.

\bibliographystyle{plain}
\bibliography{bibliography.bib}

\end{document}